\documentclass[lettersize,journal]{IEEEtran}
\usepackage{amsmath,amsfonts,amssymb}
\usepackage{algorithmic}
\usepackage{algorithm}
\usepackage{array}
\usepackage[caption=false,font=normalsize,labelfont=sf,textfont=sf]{subfig}
\usepackage{textcomp}
\usepackage{stfloats}
\usepackage{url}
\usepackage{verbatim}
\usepackage{graphicx}
\usepackage{cite}
\usepackage{multirow}
\begin{document}

\title{Trimodal Context-Aware Micro-Expression Recognition in Real-World Conversations: A Dataset and a Hierarchical Fusion Network}


\author{Junbo~Wang,
        Yan~Zhao,
        Shigang~Wang,
        Jian~Wei,
        Qiming~Zhang,
        and~Yu~Wu%
\thanks{\textit{(Corresponding author: Yan Zhao.)}}%
\thanks{Junbo Wang, Yan Zhao, Shigang Wang, and Jian Wei
are with the College of Communication Engineering, Jilin University,
Changchun 130012, China
(e-mail: [zhao\_y@jlu.edu.cn]).}%
\thanks{Qiming Zhang and Yu Wu are with Shenzhen Yinglun Tech. Ltd.,
Shenzhen, Guangdong, China.}%
\thanks{This work has been submitted to the IEEE for possible publication. Copyright may be transferred without notice, after which this version may no longer be accessible.}%
}



\maketitle

\begin{abstract}
Recognizing micro-expressions in real-world conversations is challenging, as these facial movements are brief and faint, and their interpretations are strongly dependent on the contextual circumstances of their occurrence. Most existing methods are de-contextualized, as they analyze isolated video clips while disregarding the semantic and acoustic cues in the surrounding dialogue that help explain the emergence of an expression. Decades of psychological studies have shown that social context exerts influences on facial behavior over two key dimensions: how an expression is interpreted and how strongly it is generated to begin with. However, existing micro-expression research rarely incorporates such contextual factors into its models. As an example, provocative utterances from an interlocutor may elicit suppressed negative affective reactions, which isolated silent clips are unable to account for. We propose TriCoME, a trimodal dataset consisting of visual, audio and text data sourced from real-world conversations. In this dataset, each micro-expression is paired with its concurrent utterance and preceding context. We further annotate the context based on conversational role to denote two categories: external stimuli from an interlocutor's utterance and internal background from the subject's own utterance. On this basis, we propose the Hierarchical Focusing Fusion Network (HFFNet), which aligns acoustic and semantic streams at fine granularity, compresses long dialogue sequences, modulates attention toward conversational roles via a context-type embedding, and routes each sample between visual-only and multimodal decisions based on visual confidence to prevent noisy context from degrading recognition performance. Under leave-one-subject-out cross-validation, HFFNet attains 85.89\% accuracy and 83.26\% UF1, and the ablations confirm that modeling conversational context and controlling multimodal noise are both essential to this gain.
\end{abstract}


\section{Introduction}
\IEEEPARstart{M}{icro-expressions} (MEs) are involuntary, transient facial movements of low intensity and short duration, usually shorter than 500ms, that leak out when a person tries to conceal a genuine emotion \cite{ekman2009telling,ekman1969nonverbal}. Because they betray feelings that people would rather hide, they are of lasting interest for high-stakes settings such as security screening, clinical assessment, and deception analysis \cite{li2022deep}.

Over the past decade, micro-expression recognition (MER) has moved from hand-crafted descriptors to deep spatiotemporal networks, and the ability to capture faint facial dynamics has improved considerably. Nevertheless, the great majority of this work stays within a single visual modality and treats each expression as an isolated clip. This setting is de-contextualized, and it overlooks a basic fact about social behavior: a micro-expression does not appear in a vacuum but inside a flow of interaction. When someone suppresses a real emotion during a conversation, the visible facial action is only the surface of a deeper conflict, and the cues that provoked it, such as a pointed question or a shift in the other party's tone, live in the surrounding audio and language rather than on the face itself.

The claim that context matters is not merely intuitive; it is supported by two separate lines of psychological evidence. On the interpretation side, context is what disambiguates an emotional face \cite{barrett2011context}, face and context are integrated automatically rather than optionally \cite{aviezer2011automaticity}, and for micro-expressions specifically, the same expression is judged differently once the surrounding emotional context changes \cite{zhang2014emotional}. On the production side, the evidence is stronger still. Naturalistic observation of bowlers showed that people smiled when socially engaged rather than simply when they scored well \cite{kraut1979social}. Controlled experiments demonstrated that expressive intensity is governed jointly by the underlying affective state and the social situation \cite{hess1995intensity}, and that even an implicit audience, a friend believed to be in another room, potentiates smiling in an otherwise solitary viewer \cite{fridlund1991sociality}. Recent work using automated facial tracking reports that facial and gestural expressions also vary depending on whether other people are present in socio-emotional settings \cite{heesen2024impact}, and this body of results has been synthesized into the view that facial displays function as instruments of social influence \cite{crivelli2018facial}. In short, who is present and what they have just said bear on both how a facial movement is read and whether it surfaces at all.

This dependence on context is most visible in genuine conversational settings such as social deduction games, interrogations, or counseling sessions, where the visual modality alone is often too ambiguous to reveal the intent behind a fleeting expression. The wider affective-computing community has begun to recognize this issue. Recent micro-expression challenges have moved toward question answering and long-video understanding in realistic settings \cite{fan2026megc}, which call for reasoning over context rather than over a trimmed clip. Applying this finding to build a practical conversational MER system, however, raises three problems that are still mostly unresolved. First, context is not uniform in its role. An interlocutor's utterance usually acts as an external trigger, whereas the subject's own preceding speech serves as background. These two types of information should not be treated in the same manner. Second, processing complete acoustic sequences and dialogue transcripts requires substantial computational resources. Third, adding modalities does not necessarily improve recognition. When visual features already support a confident prediction, unrelated speech or noisy text can interfere with recognition if fusion is applied indiscriminately.

\begin{figure}[!t]
\centering
\includegraphics[width=2.5in]{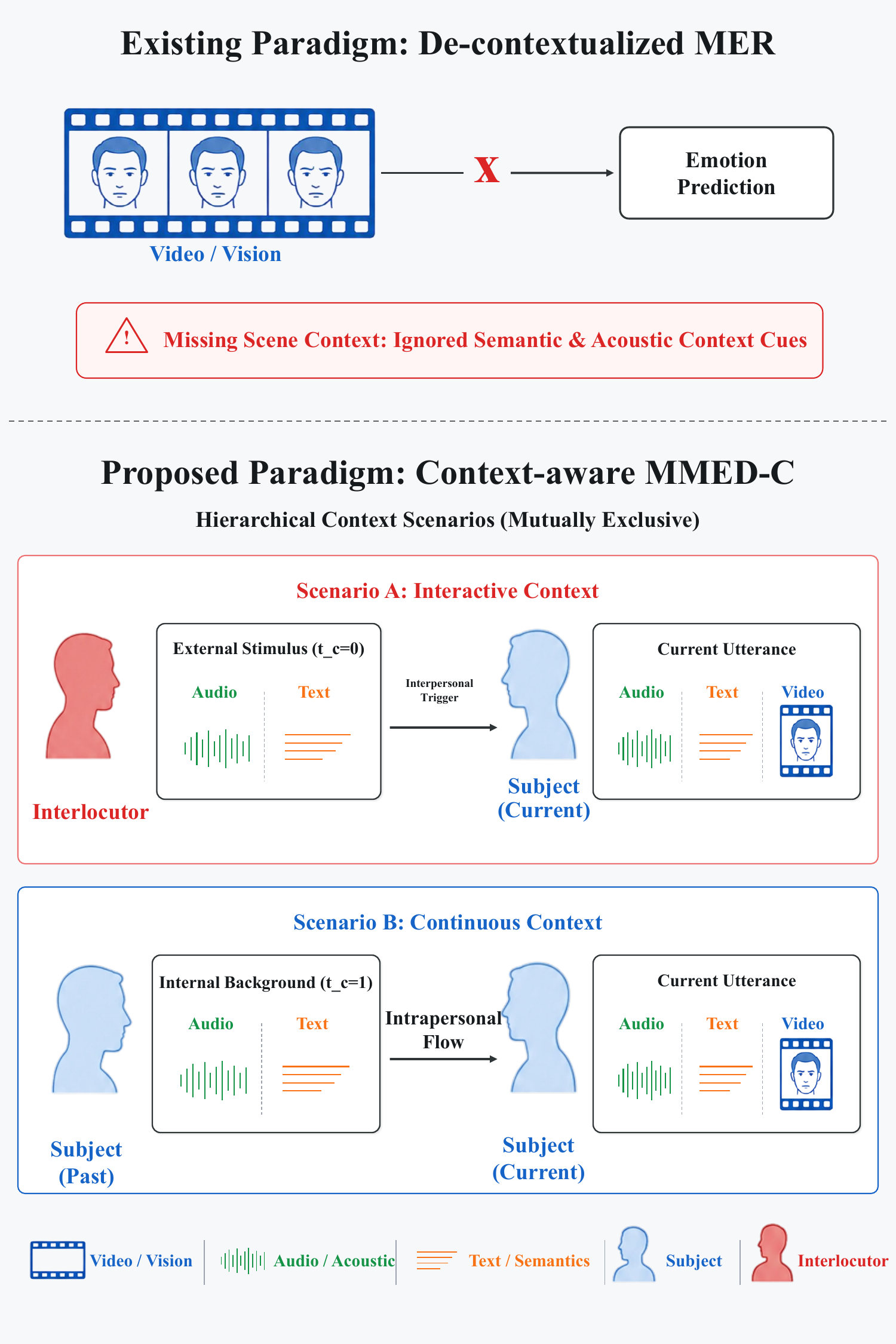}
\caption{Conceptual comparison between conventional and proposed MER paradigms.(a) De-contextualized paradigm: isolated video clips are analyzed without conversational context, ignoring semantic and acoustic elicitation cues that reveal why a micro-expression occurs.(b) Proposed context-aware TriCoME paradigm with two mutually exclusive hierarchical context scenarios: Interactive Context ($t_c=0$), where an interlocutor's utterance serves as an external stimulus triggering the subject's micro-expression (interpersonal reaction); and Continuous Context ($t_c=1$), where the subject's own preceding utterance provides internal background flow (intrapersonal coherence).}
\label{fig_4}
\end{figure}

We address these problems from both the data side and the model side. On the data side we build TriCoME, a trimodal micro-expression dataset that pairs every expression with the current utterance and the preceding context and, unlike previous de-contextualized resources, records the conversational role of that context, as illustrated in Fig.~\ref{fig_4}. This gives a basis for studying how a micro-expression is elicited rather than only how it looks. On the model side we propose the Hierarchical Focusing Fusion Network (HFFNet), which is organized around the way a human listener allocates attention. Instead of concatenating modalities in a single step, HFFNet lets verbal content and vocal tone interact at a fine granularity so that momentary mismatches between the two are preserved, encodes the asymmetric roles of dialogue participants through a context-type embedding, and decides for each sample whether contextual assistance is actually needed. This last step is governed by an adaptive routing mechanism that consults the confidence of the visual branch and brings in the linguistic context only when the visual evidence is uncertain, which mirrors how people seek extra cues under doubt while keeping computation in check.

In summary, the main contributions of this paper are as follows:

1)We build TriCoME, a trimodal micro-expression dataset for real conversations. Beyond synchronized video and audio, it provides manually transcribed text and, to our knowledge, the first context annotations in this field that distinguish interpersonal triggers from the subject's own statements. The resource offers a practical basis for studying conversational micro-expressions and for analyzing how the surrounding speech relates to their onset.

2)We propose HFFNet, a hierarchical fusion framework. To process long dialogue histories without overfitting, we pair a Sequence Compressor with a Fine-grained Cross-modal Interaction module. The former condenses a variable-length fused sequence into a fixed-length representation, which lowers the quadratic cost of self-attention from $O(L^2)$ to $O(L_{comp}^2)$ with $L_{comp}=256$. The latter preserves the frame-level correspondence between acoustic and semantic streams that a global pooling step would otherwise erase.

3)We design a Contextual Transformer with a Context Type Embedding and an Adaptive Multi-modal Routing strategy. By distinguishing dialogue roles and consulting audio and text when visual evidence is uncertain, the framework reduces the risk that irrelevant context leads to an incorrect prediction.

4)Extensive leave-one-subject-out experiments on TriCoME show that HFFNet reaches 85.89\% accuracy and 83.26\% UF1, and a series of ablations isolates the contribution of each component.

\section{Related Work}
We review prior work along four lines that bear on conversational micro-expression recognition: visual MER, multimodal fusion, context-aware affective computing, and benchmark datasets. Each line outlines a limitation that this paper seeks to address.

\subsection{Visual MER}
As one of the most challenging tasks in affective computing, visual Micro-Expression Recognition (MER) has undergone long-term technological iteration. Early research relied on expert-knowledge-driven hand-crafted features, such as LBP-TOP\cite{zhao2007dynamic} and its variants, attempting to capture subtle facial motions by constructing compact texture descriptors. Subsequently, the deep learning paradigm became dominant, with 3D-CNNs\cite{reddy2019spontaneous} and optical-flow-based two-stream networks\cite{liong2018less} significantly enhancing feature discriminability.

More recently, to cope with limited training data, the field has turned to Transformer-based modeling and self-supervised pre-training. Micron-BERT \cite{nguyen2023micron} dispenses with optical flow and learns subtle movements directly from raw pixels through a diagonal micro-attention design. PrefAce \cite{hu2024preface} uses face-structure-aware self-supervised pre-training to sharpen sensitivity to small facial deformations. Working from RGB alone, the motion-magnification network of Wu et al. \cite{wu2024micro} couples amplification of weak motion with attention modules to strengthen the representation of local pixel changes.

Despite the effectiveness of uni-modal visual methods in detecting subtle pixel-level motion, their analysis is primarily grounded in signal processing. While adept at describing facial actions, such approaches may not fully elucidate the underlying intent or causality due to the lack of multimodal context. Consequently, their generalization capability can be constrained when addressing the high ambiguity inherent in real-world conversational scenarios.

\subsection{Multi-modal Exploration}
To move beyond a single modality, researchers have explored two main fusion strategies.

The first strategy is intra-visual fusion, which enriches the representation with auxiliary visual cues such as optical flow, depth, or action units. Graph-AU \cite{lei2021micro} builds a facial graph and injects action units as prior knowledge into feature extraction, and the two-stream adaptive multi-attention graph network of Kumar and Bhanu \cite{kumar2024uncovering} combines spatial node features with optical-flow dynamics. Despite the multimodal label, these are augmentations internal to the visual channel and bring in no non-visual context.

The second strategy turns to physiological signals. Zhao et al. \cite{zhao2024micro}, for example, examined how EEG functional networks change while a micro-expression occurs. This opens a window on the underlying mechanism, but the intrusive nature of EEG acquisition makes it not feasible in natural interaction.

In the above strategies the auxiliary signal is either a re-encoding of the face itself or an internal bodily measurement. Neither brings in the audio and the language of the conversation in which the expression is embedded, so the elicitation cues carried by speech remain outside the model.

\subsection{Context-aware Affective Computing Paradigm}
Robust conversational MER needs more than a stronger visual model, whether that strength comes from internal augmentation or physiological signals. The context governs how an emotional face is interpreted \cite{barrett2011context,aviezer2011automaticity} and also how strongly it is expressed \cite{hess1995intensity,fridlund1991sociality,crivelli2018facial}. The surrounding conversation is therefore part of the signal rather than a nuisance variable.

In macro-emotion recognition, and in Emotion Recognition in Conversation (ERC) in particular, modeling context is now standard. From DialogueRNN \cite{majumder2019dialoguernn} to COSMIC \cite{ghosal-etal-2020-cosmic}, which brings in commonsense knowledge, researchers have studied how to capture the dependencies that run through a conversation. GraphCFC \cite{li2023graphcfc} uses directed graph networks to model speaker dependency and temporal context, and Yang et al. \cite{yang2022contextual} propose a hierarchical interaction framework that couples contextual interaction with cross-modal alignment for speech emotion recognition. The paradigm has also been combined with multimodal large language models: MicroEmo \cite{zhang2024microemo} attends to local facial dynamics alongside utterance-level context over video, audio, and transcribed speech. It is aimed at open-ended emotion description on macro-level clips rather than expert-annotated micro-expressions, however, and it treats context as an undifferentiated time window.

We borrow from these macro-level paradigms and use acoustic and semantic information to support micro-expression understanding, but a direct transfer is not enough. ERC models mainly track how an emotional state evolves and stays consistent across a long dialogue, whereas a micro-expression is an instantaneous leakage triggered by a specific moment, and its relation to context is functionally asymmetric. An interlocutor's utterance, such as a rebuttal or a probing question, often acts as an external trigger that induces suppression, while the subject's own preceding statement is closer to an internal background that maintains coherence. ERC methods model speaker dependency yet typically fail to differentiate between context as an active stimulus and a passive background.

\subsection{Micro-expression Datasets}
High-quality datasets have driven progress in micro-expression research. Early benchmarks such as SMIC \cite{pfister2011recognising,6553717}, CASME II \cite{yan2014casme}, and SAMM \cite{davison2016samm} set up the basic protocols for MER and remain influential. However, these datasets were recorded in tightly controlled laboratories under a passive-viewing paradigm, which does not reflect the active nature of real-world interaction.

To improve ecological validity, later datasets such as MMEW \cite{ben2021video}, CAS(ME)$^3$ \cite{li2022cas}, 4DME \cite{li20224dme}, and DFME \cite{10354506} enlarged the scale and the range of modalities. CAS(ME)$^3$ \cite{li2022cas} added ambient audio and physiological signals under a mock-crime interrogation paradigm, and the in-the-wild NaME dataset \cite{liu2025name} gathered spontaneous micro-expressions from real-world online videos to move beyond the laboratory. That said, existing datasets suffer from a key limitation: few are equipped to provide transcribed dialogue text alongside context annotations for distinguishing conversational roles, such as interlocutor trigger and self-statement background.

Visual MER primarily relies on facial motion, and its auxiliary cues often remain within the visual modality or require intrusive physiological measurements. Context-aware affective models do not usually distinguish between context as a possible external stimulus and as the speaker's own background, while available ME datasets rarely pair audio and transcripts with annotations of these roles. To address these limitations, we introduce TriCoME with synchronized video, audio, manually transcribed text, and role-labeled context, and propose HFFNet to integrate these sources while limiting interference from irrelevant context.

\section{Methodology}
This section introduces the TriCoME dataset and the proposed Hierarchical Focusing Fusion Network (HFFNet).

\begin{figure*}[!t]
\centering
\captionsetup[subfloat]{font=footnotesize}
\subfloat[]{\includegraphics[width=2.5in]{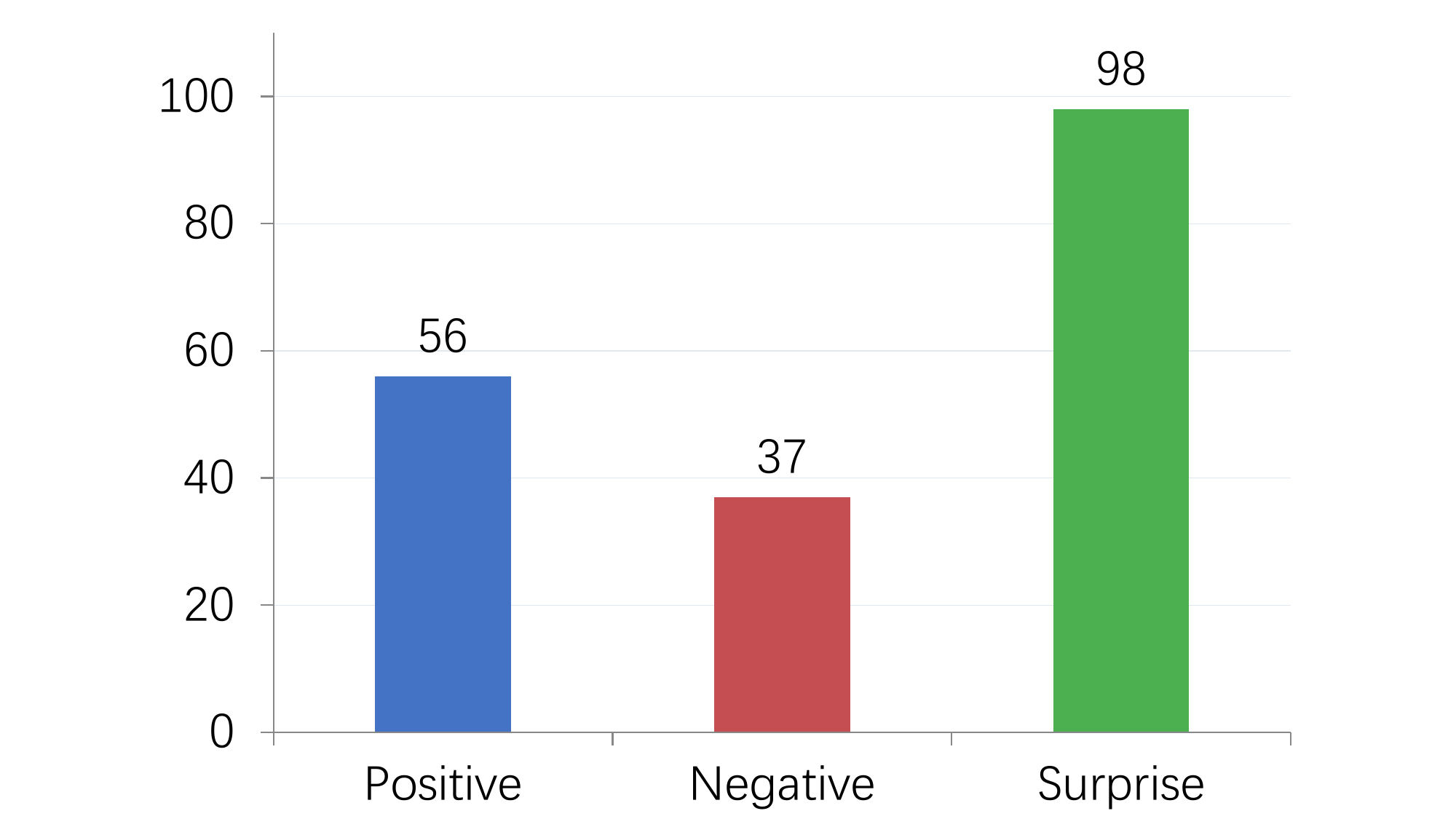}%
\label{fig_first_case}}
\hfil
\subfloat[]{\includegraphics[width=2.5in]{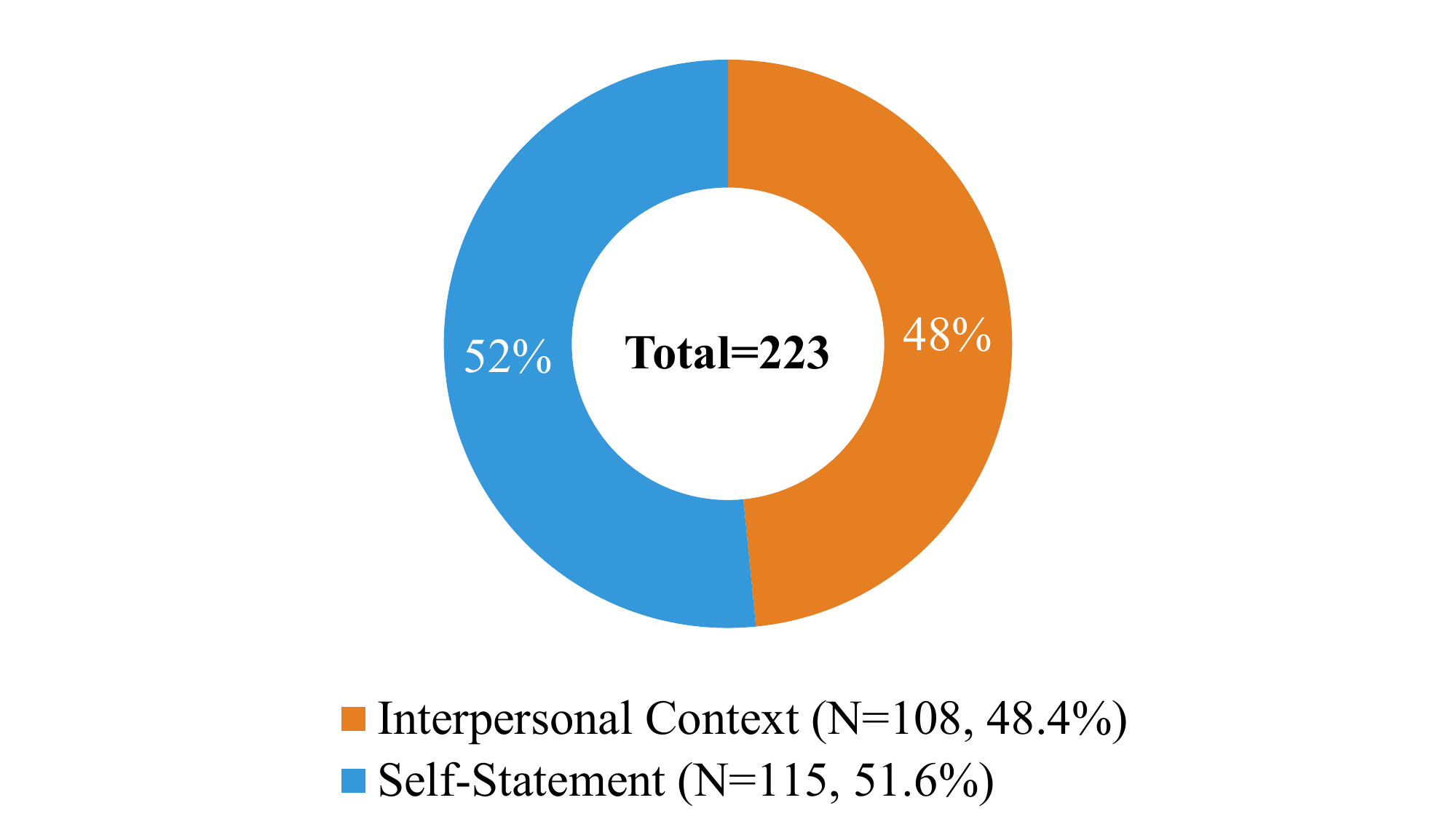}%
\label{fig_second_case}}
\caption{Statistical overview of the TriCoME dataset. (a) Emotion category distribution across 191 micro-expression samples. (b) Distribution of context audio segments (N=223), where the 108 interpersonal segments are derived from 76 samples with multi-segment contexts.}
\label{fig_sim}
\end{figure*}

\subsection{TriCoME Dataset}
TriCoME is a trimodal, in-the-wild micro-expression dataset built to address the lack of transcribed dialogue and role-level context annotations in existing ME datasets. Each micro-expression is paired with the audio and transcript of its current utterance and with the preceding conversational context. The context is labeled as either another speaker's utterance or the subject's own preceding speech.

\textbf{Data source.} Following the in-the-wild practice of MEVIEW \cite{husak2017spotting}, we source material from publicly available recordings rather than a laboratory. Specifically, we use publicly available videos of seven online Werewolf matches. Werewolf is a social deduction game in which players conceal their true roles and intentions under time pressure, which naturally provokes suppression attempts and the micro-expressions that leak through them, while the spoken turns of the game supply a continuous stream of audio and language. The recordings are ordinary web video and therefore carry the native frame rate of the source, about 30 frames per second, rather than the 100 to 200 frames per second of high-speed laboratory capture. This is a known trade-off of in-the-wild data \cite{husak2017spotting}, and, as described below, we compensate for it by annotating every expression frame by frame.

\textbf{Annotation protocol.} Annotators certified in the Facial Action Coding System (FACS) labeled the data in stages. Candidate segments were first screened, and a certified coder then inspected each one frame by frame to mark the onset, apex, and offset, following the common criterion that the whole event lasts no more than 500ms; action units were coded and an emotion label was assigned to every validated expression. The full set was independently annotated a second time by another certified coder, and the agreement over action units, computed as $2|A_1 \cap A_2| / (|A_1| + |A_2|)$, reached 0.89, and the remaining disagreements are resolved through discussion. Emotion categories follow the mapping used in 4DME \cite{li20224dme}, with the repression category excluded. In these conversational clips, mouth opening (AU50) frequently accompanies speech. This makes the temporal criteria used to label repression difficult to apply consistently, so we do not assign that category.

\textbf{Modalities and context.} Using the temporal marks, we cut the complete sentence spoken at the moment of onset and treat it as the current utterance. We then look back and cut the turn immediately preceding it as the context utterance, and we label the context by its conversational role: an interlocutor utterance, that is speech from another player directed at the subject and acting as an external stimulus, or a self-statement, that is the subject's own preceding speech acting as internal background. All audio was transcribed manually, which yields the paired text and produces the trimodal video, audio, and text organization of TriCoME.

\textbf{Statistics.} The recognition dataset contains 191 micro-expression samples in three emotion categories: 56 positive, 37 negative, and 98 surprise, as shown in Fig.~\ref{fig_sim}. Eight further samples fall into an others category; we keep them in the released annotations for completeness but exclude them from the recognition experiments in this paper. Each expression has one current-utterance audio segment, for 191 in total. For context we collected 223 segments, comprising 115 self-statement segments, one per sample, and 108 interlocutor segments drawn from 76 samples in which some turns span more than one exchange. Manual transcription produces one text segment per audio segment.

Table~\ref{tab:dataset_comparison} compares TriCoME with existing spontaneous ME datasets in terms of acquisition setting, modalities, transcribed text, and context-role annotations.

\begin{table*}[!t]
\caption{Comparison of TriCoME with Existing Spontaneous Micro-Expression Datasets\label{tab:dataset_comparison}}
\centering
\footnotesize
\setlength{\tabcolsep}{2.5pt}
\begin{tabular}{lccccclcc}
\hline
\multirow{2}{*}{Dataset} & \multirow{2}{*}{Year} & \multirow{2}{*}{Subj} & \multirow{2}{*}{Samples} & \multirow{2}{*}{FPS} & \multirow{2}{*}{Modalities} & \multirow{2}{*}{Elicitation} & \multirow{2}{*}{Context} & \multirow{2}{*}{Text} \\
&&&&&&&&\\
\hline
SMIC \cite{6553717} & 2013 & 16 & 164 & 100 & V & Passive & $\times$ & $\times$ \\
CASME \cite{yan2013casme} & 2013 & 19 & 195 & 60 & V & Passive & $\times$ & $\times$ \\
CASME II \cite{yan2014casme} & 2014 & 26 & 247 & 200 & V & Passive & $\times$ & $\times$ \\
SAMM \cite{davison2016samm} & 2016 & 32 & 159 & 200 & V & Passive & $\times$ & $\times$ \\
MEVIEW \cite{husak2017spotting} & 2017 & 16 & 29 & 30 & V & In-the-wild & $\times$ & $\times$ \\
CAS(ME)$^2$ \cite{qu2017cas} & 2018 & 22 & 57 & 30 & V & Passive & $\times$ & $\times$ \\
MMEW \cite{ben2021video} & 2021 & 36 & 300 & 90 & V & Passive & $\times$ & $\times$ \\
CAS(ME)$^3$ \cite{li2022cas} & 2022 & 247 & 1,109 & 30 & V+D+P+A & Interaction$^*$ & $\times$ & $\times$ \\
4DME \cite{li20224dme} & 2022 & 65 & 1,068 & 30-60 & V+D+4D & Passive & $\times$ & $\times$ \\
DFME \cite{10354506} & 2023 & 671 & 7,526 & 200-500 & V & Passive & $\times$ & $\times$ \\
MMME \cite{ma2025mmme} & 2025 & 75 & 634 & 150 & V+P & Passive & $\times$ & $\times$ \\
NaME \cite{liu2025name} & 2025 & 106 & 506 & -- & V & In-the-wild & $\times$ & $\times$ \\
\textbf{TriCoME (Ours)} & 2026 & \textbf{16} & \textbf{191} & \textbf{30$^\dagger$} & \textbf{V+A+T} & \textbf{Interaction} & \textbf{\checkmark} & \textbf{\checkmark} \\
\hline
\end{tabular}

\vspace{2pt}
\begin{minipage}{\textwidth}
\footnotesize
\textbf{Note:} V = Video, A = Audio, T = Transcribed Text, D = Depth, P = Physiological signals. The Context column indicates whether a dataset annotates the conversational role of context, separating interpersonal triggers from self-statement background. The Text column indicates manually transcribed dialogue text. $^*$CAS(ME)$^3$ uses a mock-crime interrogation paradigm; its audio is an ambient recording without transcribed text or role-level context. $^\dagger$TriCoME is sourced from publicly available online videos and keeps the native frame rate of the web video; onset, apex, and offset are localized by frame-by-frame inspection.
\end{minipage}
\end{table*}

\textbf{Ethics and release.} The source material consists of publicly available online videos. We do not recruit participants or record new videos for this study. To respect the rights of the original uploaders and the privacy of the players, we do not redistribute any edited video. The release consists of the annotation file, which holds the temporal marks, action units, emotion labels, context-role labels, and transcripts, together with the URLs of the source recordings so that the visual frames can be located from the public originals. Access is granted on request: interested researchers sign a usage agreement, and the annotation file is provided after review.

\subsection{Overall Architecture of HFFNet}
\begin{figure*}[!t]
\centering
\includegraphics[width=0.95\textwidth]{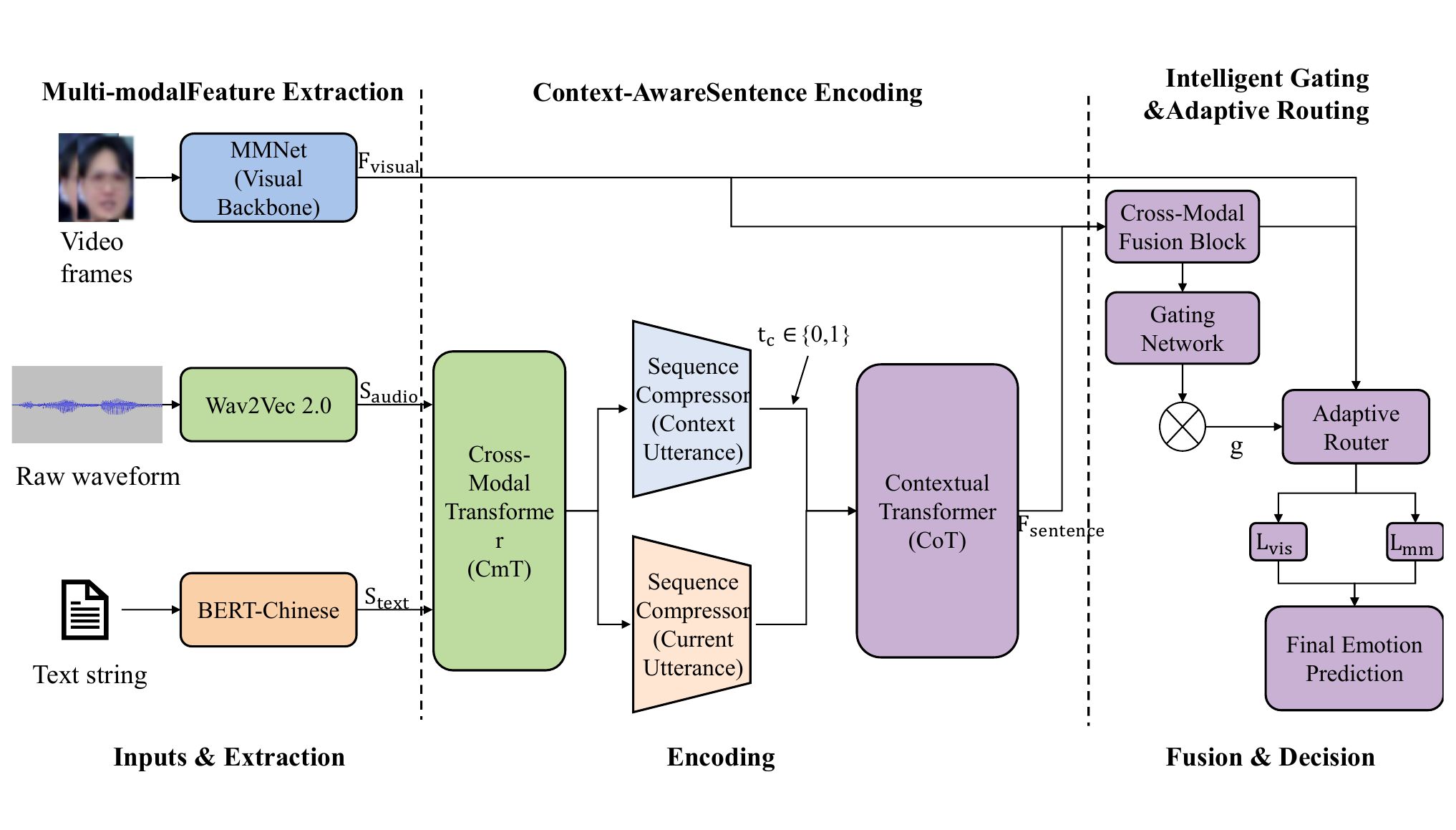}
\caption{Architecture of the proposed Hierarchical Focusing Fusion Network (HFFNet) for context-aware micro-expression recognition. The pipeline comprises three stages. (1) Multi-modal Feature Extraction: MMNet extracts visual features $F_{visual}$ from video frames; Wav2Vec 2.0 with a sliding-window strategy extracts acoustic features $S_{audio}$ from raw waveforms; BERT extracts semantic features $S_{text}$ from transcribed text. (2) Context-Aware Sentence Encoding: A Cross-Modal Transformer (CmT) fuses $S_{audio}$ and $S_{text}$ through fine-grained interaction, followed by a Sequence Compressor that condenses the fused sequence into a compact representation; a Contextual Transformer (CoT) then injects context type embeddings $t_{c} \in \{0,1\}$ to distinguish interlocutor utterances (external triggers) from self-statements (internal background), yielding the sentence feature $F_{sentence}$. (3) Intelligent Gating \& Adaptive Routing: $F_{visual}$ and $F_{sentence}$ interact via a cross-modal fusion block and an enhanced gating network with element-wise gate vector $g$; an adaptive router dynamically selects between the visual-only prediction $L_{vis}$ and the multimodal prediction $L_{mm}$ based on visual confidence, producing the final emotion classification.}
\label{fig_2}
\end{figure*}

Fig.~\ref{fig_2} presents HFFNet. The network contains a multimodal feature extractor, a context-aware sentence encoder, and a fusion and routing module. The feature extractor represents facial motion, speech, and transcript content using a task-adapted MMNet visual encoder, Wav2Vec 2.0, and BERT, respectively. Within the sentence encoder, a Cross-Modal Transformer first combines audio and text, and a Sequence Compressor reduces the length of the fused sequences. A Contextual Transformer then uses the preceding utterance and its role label to condition the representation of the current utterance. Finally, cross-modal attention and an element-wise gate combine the visual and sentence features. The adaptive router selects a visual-only or multimodal prediction according to visual confidence.

\subsection{Multi-modal Feature Extraction}
This section describes feature extraction for the visual, audio, and text modalities. We use a task-adapted MMNet for vision and pretrained Wav2Vec 2.0 and BERT models for audio and text, respectively.

1) Visual Feature Extraction: We use a task-adapted implementation of MMNet \cite{ijcai2022p150} as the visual backbone of HFFNet, based on the comparison in Table~\ref{tab:visual_baseline}. The implementation retains MMNet's continuous attention (CA) and position calibration (PC) mechanisms. Given preprocessed onset and apex images, the motion branch processes $I_{apex}-I_{onset}$ through convolutional layers and four CA blocks, while the PC branch encodes the onset image resized to $14\times14$. The two $512\times14\times14$ feature maps are added element-wise and flattened into $f_v\in\mathbb{R}^{100352}$. For the three-class task on TriCoME, the visual classifier applies Dropout with probability $0.5$ followed by a bias-free linear layer with three outputs. A separate feature interface exposes $f_v$ for HFFNet, whose visual projection maps it to the fusion representation $F_{visual}\in\mathbb{R}^{d}$ with $d=256$. These adaptations concern the task-specific classifier and the feature and prediction interfaces required for fusion and routing; the CA and PC computations are inherited from MMNet.

2) Audio Feature Extraction: To capture subtle paralinguistic cues from raw waveforms, we utilize the pre-trained Wav2Vec 2.0 Base model\cite{baevski2020wav2vec} as the acoustic feature extractor. This model, optimized via self-supervised contrastive learning on massive unlabeled speech data, extracts robust context-dependent acoustic representations. In order to address the issue of ultra-long speech segments common in conversational scenarios, which may lead to GPU memory overflow, we design a Sliding Window Chunking Strategy. Specifically, we segment long audio into continuous chunks of 10 seconds with a 50\% overlap. Each chunk is independently fed into the Wav2Vec2 model to extract the last hidden state. To preserve complete temporal dynamics, we discard traditional global average pooling and instead concatenate the features of all chunks along the time dimension, forming a complete acoustic feature sequence $S_{audio} \in \mathbb{R}^{T_a \times 768}$, where $T_a$ denotes the total number of frames. Furthermore, to prevent overfitting, the pre-trained parameters of Wav2Vec2 are frozen.

3) Text Feature Extraction: For dialogue transcripts, we utilize the pre-trained BERT-Base-Chinese model\cite{devlin2019bert} to extract deep semantic features. Similar to audio processing, we employ a sliding window mechanism (maximum length 128 Token, overlapping 32 Token) to handle long texts.A necessary implementation detail concerns padding. In standard chunking processing, chunks falling short of the required length are padded with [PAD] tokens. If these meaningless zero vectors are directly concatenated, they dilute the semantic density of the sequence. Therefore, we explicitly remove all feature vectors corresponding to [PAD] before concatenation, retaining only the embeddings of real tokens. Ultimately, effective features from all chunks are concatenated into a complete semantic sequence $S_{text} \in \mathbb{R}^{L_t \times 768}$, where $L_t$ is the total number of effective tokens.

\subsection{Context-Aware Sentence Encoding}
Emotions in dialogue are not only dependent on current visual manifestations but also conveyed through verbal content (text) and vocal tone (audio). Moreover, they are deeply influenced by conversational context. Inspired by the hierarchical interaction approach in \cite{yang2022contextual}, we implement critical adaptive improvements tailored to micro-expression characteristics. Unlike general conversational emotion models, we explicitly categorize context based on interaction nature into two types: 1) Interlocutor's Utterance ($t_c=0$), serving as an external stimulus source; and 2) Self-Statement ($t_c=1$), serving as an internal background monologue.

Our encoder adopts an interact-then-modulate strategy: we first fuse audio and text separately for the context and for the current utterance to obtain complete multimodal representations, and a Contextual Transformer then uses the fused context as a prior to modulate the understanding of the current utterance. In this way the model captures not only what was said but also who it was said to.

\begin{figure*}[!t]
\centering
\includegraphics[width=0.95\textwidth]{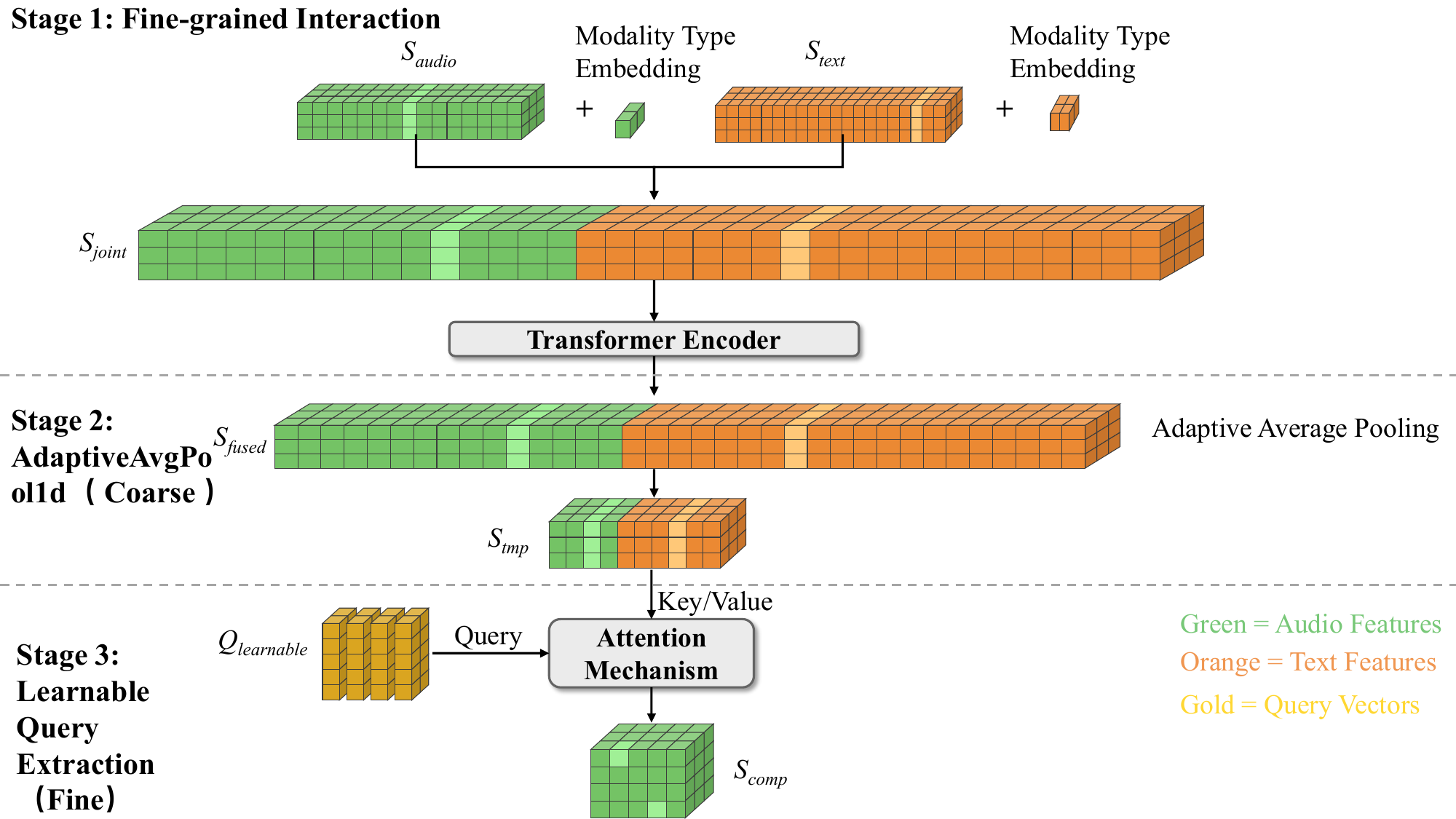}
\caption{Fine-grained cross-modal interaction and two-stage hybrid sequence compression pipeline (the interact phase of our encoder). For visual clarity, $S_{audio}$ and $S_{text}$ are depicted after linear projection to $d=256$ (the projection step is omitted from the diagram). Stage~1: learnable modality type embeddings are added to the projected features; the two sequences are concatenated into $S_{joint}$ and processed by a Transformer encoder to achieve token-level audio-text interaction. Stage~2: adaptive average pooling rapidly compresses the fused sequence $S_{fused}$ to an intermediate representation $S_{tmp}$, eliminating temporal redundancy. Stage~3: a lightweight multi-head attention mechanism with learnable queries $Q_{learnable}$ retrieves and reconstructs the most discriminative semantics from $S_{tmp}$, yielding a compact fixed-length output $S_{comp}$ ($L_{comp}=256$).}
\label{fig_5}
\end{figure*}

1) Cross-Modal Transformer: To bridge the heterogeneity between audio and text and capture elicitation cues for micro-expressions, we build fine-grained cross-modal interaction into CmT and adopt an interact-then-summarize strategy. Unlike methods that pool each modality into a global vector before fusion, our design rests on a cognitive observation: a micro-expression is often triggered by an instantaneous mismatch between verbal content and paralinguistic cues, for example saying that one is fine while the voice trembles \cite{ekman1969nonverbal}. If global pooling is performed too early, this fleeting frame-level conflict information is smoothed out. Therefore, we preserve the complete sequence structures of $S_{audio}$ and $S_{text}$. As illustrated in Fig.~\ref{fig_5} (Stage 1), we first project both sequences to a common dimension $d=256$ via linear layers, then add learnable modality type embeddings to the projected features. The two projected sequences are subsequently concatenated along the sequence dimension to form a joint sequence $S_{joint}$. When the diagram depicts $S_{audio}$ and $S_{text}$, they correspond to these 256-dimensional projected features rather than the original 768-dimensional raw features. The Cross-Modal Transformer then utilizes multi-layer Transformer Encoders to perform deep interaction across the full sequence, where the self-attention mechanism allows every token in the text to attend directly to the corresponding acoustic frames in the audio stream. This aligns semantics with tone at a fine granularity and preserves the contextual cues needed to judge insincerity.

2) Sequence Compressor: Since the fused sequence output by Cross-Modal Transformer can encompass thousands of time steps, feeding it directly into subsequent modules would incur prohibitive computational complexity ($O(L^2)$) and potential memory overflow. To address this, we introduce a lightweight Sequence Compressor to eliminate temporal redundancy in high-dimensional fused features. As shown in Fig.~\ref{fig_5} (Stages 2--3), this module employs a two-stage coarse-to-fine compression strategy. First, Adaptive Average Pooling is used to rapidly reduce the long sequence to an intermediate length without parameters, providing a global context overview while significantly lowering computational load. Second, to prevent the loss of critical details caused by pooling, we introduce a Lightweight Multi-head Attention mechanism based on Learnable Queries. These queries retrieve and reconstruct the semantic information most critical for emotion recognition from the coarse features, ultimately generating a compact feature sequence of fixed length ($L_{comp}=256$).This design maintains information density while effectively mitigating the overfitting risks associated with long sequences. Formally, the compression process is expressed as:

\begin{equation}
\label{deqn_ex1a}
S_{tmp} = \text{AdaptivePool}(S_{fused})
\end{equation}

\begin{equation}
\label{deqn_ex1a}
S_{comp} = \text{Attention}(Q_{learnable}, S_{tmp}, S_{tmp})
\end{equation}
where $S_{fused}$ denotes the high-dimensional long fused sequence output by the Cross-Modal Transformer, and $S_{tmp}$ represents the intermediate features obtained after dimensionality reduction via adaptive pooling (serving simultaneously as Key $K$ and Value $V$ in the attention mechanism). $Q_{learnable}$ refers to a set of learnable query vectors designed to reconstruct critical semantics (acting as Query $Q$), while $S_{comp}$ denotes the final output, which is a fixed-length compressed feature sequence.

3) Contextual Transformer: Conversational context can come from another speaker or from the target speaker's own preceding utterance. These sources play different roles in interpreting the current facial movement, so we design the Contextual Transformer to represent them explicitly. In Interactive Context ($t_c=0$), another speaker's question or rebuttal provides a possible external stimulus. In Continuous Context ($t_c=1$), the target speaker's preceding speech provides background for the current utterance. This representation can also be constructed for other conversational datasets that provide paired video, audio, transcripts, and speaker-turn annotations. After sequence compression, a Context Type Embedding identifies the role of the context. We map the label $t_c \in \{0,1\}$ to a type vector through $\mathcal{E}: \{0,1\} \rightarrow \mathbb{R}^{d}$ and add this vector to every token in the compressed context features $H_{ctx}^{comp}$:

\begin{equation}
\label{deqn_ex1a}
H_{ctx}^{typed} = H_{ctx}^{comp} + \mathcal{E}(t_c)
\end{equation}
The added type vector tells the Transformer whether the context comes from another speaker or from the target speaker. Positional encoding identifies where a token occurs in a sequence, whereas this embedding identifies the conversational role of the context. The model can therefore condition its interpretation of the current utterance on either a possible external stimulus ($t_c=0$) or the speaker's own preceding statement ($t_c=1$).
After injecting type information, Contextual Transformer employs a sequence-level Cross-Attention mechanism to modulate semantics. We treat the feature sequence of the current utterance, $H_{curr}$, as the Query, and the concatenation of the typed context sequence and current utterance, $[H_{ctx}^{typed}; H_{curr}]$, as Key and Value. This enables the current utterance to precisely aggregate relevant historical cues with explicit awareness of whether it is responding to another or continuing itself. The CoT output is finally processed via attention pooling to generate the context-aware sentence feature vector $F_{sentence}$.

\subsection{Intelligent Gating Fusion and Classification}
Upon obtaining the visual feature $F_{visual}$ and the context-aware sentence feature $F_{sentence}$, we employ an Intelligent Gating Fusion Module (Final Fusion Module) for multi-modal integration.

To further promote deep synergy between modalities, we first introduce a Cross-Modal Fusion Block before the gating mechanism. This block utilizes a bi-directional attention mechanism to enable pre-interaction between visual features and sentence semantic features, thereby enhancing feature complementarity.

Formally, given the visual feature $F_{\text{visual}}$ and the context-aware sentence feature $F_{\text{sentence}}$, the fusion process is:

\begin{equation}
\label{eq:visual_update}
\begin{split}
F_{\text{visual}}' = \text{LayerNorm}\bigl( F_{\text{visual}} + \text{CrossAttn}(
      & Q{=}F_{\text{visual}}, \\
      & K{=}F_{\text{sentence}}, \\
      & V{=}F_{\text{sentence}} ) \bigr)
\end{split}
\end{equation}

\begin{equation}
\label{eq:sentence_update}
\begin{split}
F_{\text{sentence}}' = \text{LayerNorm}\bigl( F_{\text{sentence}} + \text{CrossAttn}(
      & Q{=}F_{\text{sentence}}, \\
      & K{=}F_{\text{visual}}, \\
      & V{=}F_{\text{visual}} ) \bigr)
\end{split}
\end{equation}
where $\text{CrossAttn}(\cdot)$ denotes the multi-head cross-attention operation, and the enhanced features $F_{\text{visual}}'$ and $F_{\text{sentence}}'$ replace their original counterparts in the subsequent gating and classification steps.

Subsequently, considering the inconsistency of audio-visual information (e.g., poor video quality or ambiguous speech), we introduce an Enhanced Gating Network to dynamically regulate feature weights. The gating network consists of a 3-layer MLP taking the concatenation of the enhanced visual and sentence features as input, outputting a gating vector $g \in (0,1)^{d}$:

\begin{equation}
\label{deqn_ex1a}
g = \sigma(\text{MLP}([F_{visual}'; F_{sentence}']))
\end{equation}
where $\sigma$ is the Sigmoid activation function. This gate operates element-wise, allowing for independent weight regulation of each feature dimension.

To encourage explicit gating decisions, we introduce Binarization Decision Regularization during training. This regularization term forces the gating values to polarize by minimizing $\mathcal{L}_{reg}$, thereby avoiding ambiguous intermediate states:

\begin{equation}
\label{deqn_ex1a}
\mathcal{L}_{reg} = \lambda \cdot \text{mean}(g \odot (1-g))
\end{equation}
where $\lambda$ denotes the gating regularization weight. The final fused feature $F_{final}$ is computed as follows:

\begin{equation}
\label{deqn_ex1a}
F_{final} = [F_{visual}'; g \odot F_{sentence}']
\end{equation}
where $\odot$ denotes element-wise multiplication and $[;]$ denotes concatenation. This mechanism allows the model to suppress irrelevant or noisy context information when $g \to 0$ and fully leverage high-quality context cues when $g \to 1$.

The fused features are fed into a classifier (comprising fully connected layers, ReLU activation, and Dropout) to yield the prediction distribution of the multi-modal branch, $L_{mm}$.

\subsection{Adaptive Multi-modal Routing}
To further enhance model robustness in complex scenarios, we introduce a sample-level Adaptive Router. This mechanism addresses the negative transfer problem in multi-modal fusion, where forced fusion degrades performance if the semantic information (audio or text) is noisy or misleading.

The router dynamically determines the final prediction output based on the visual model's confidence, $c_{vis}$, defined as the maximum softmax probability of the visual-only branch:

\begin{equation}
\label{deqn_ex1a}
c_{\text{vis}} = \max\left(\operatorname{softmax}(L_{\text{vis}})\right)
\end{equation}
For samples with high confidence ($c_{vis} \ge \tau$), we deem the visual features sufficiently discriminative and directly adopt the prediction from the visual branch to avoid overfitting. For hard samples with low confidence ($c_{vis} < \tau$), we utilize the prediction from the multi-modal branch, which incorporates context information.
In the training phase, to ensure differentiability, we employ a soft routing strategy, calculating the weight of the multi-modal branch $w_{mm}$ via a Sigmoid function:

\begin{equation}
\label{deqn_ex1a}
w_{mm} = \text{sigmoid}((\tau - c_{vis}) \times T)
\end{equation}
where $\tau$ is the routing threshold, and $T$ is the temperature parameter (controlling the steepness of the Sigmoid).

\begin{equation}
\label{deqn_ex1a}
L_{final} = (1 - w_{mm}) \cdot L_{vis} + w_{mm} \cdot L_{mm}
\end{equation}
where $L_{vis}$ and $L_{mm}$ denote the prediction logits output by the visual branch and the multi-modal branch, respectively, and $L_{final}$ represents the weighted fused logits used for loss calculation. Therefore, during the training phase, instead of making a hard binary decision, the model performs linear interpolation in the decision space of both branches via the dynamic weight $w_{mm}$, thereby facilitating end-to-end differentiable optimization of the routing strategy.In the testing phase, we employ a hard routing strategy for discrete decision-making to obtain specific inference results.

\section{Experiments}
\subsection{Experimental Setup}
We use Leave-One-Subject-Out Cross-Validation (LOSO-CV) with 16 folds. In each fold, all samples from one subject form the test set, and samples from the other subjects are used for training. We report Accuracy (Acc) and Unweighted F1-score (UF1). Training hyperparameters are method-specific for the visual-backbone comparison in Table~\ref{tab:visual_baseline}; the settings for the reproduced baselines and the MMNet adaptation are specified below.

UF1 is the macro-averaged F1 used in micro-expression recognition \cite{wu2024micro}. For each evaluated emotion class $c$, its F1-score combines precision and recall:
\begin{equation}
\label{eq:uf1_definition}
\begin{aligned}
F1_c &= \frac{2TP_c}{2TP_c+FP_c+FN_c}, \\
\mathrm{UF1} &= \frac{1}{C}\sum_{c=1}^{C}F1_c,
\end{aligned}
\end{equation}
where $TP_c$, $FP_c$, and $FN_c$ are the true-positive, false-positive, and false-negative counts for class $c$, and $C$ is the number of classes included in the evaluation. ``Unweighted'' means that the class-wise F1-scores contribute equally to this average. In contrast, support-weighted F1 gives a class more weight when it contains more test samples. UF1 therefore prevents a frequent class from dominating the average solely because of its size. These averaging weights belong to the evaluation metric and do not specify the class weights used during training.

\subsection{Visual Backbone Selection}
To select a visual backbone for HFFNet, we evaluate AlexNet \cite{tricome_krizhevsky2012imagenet}, KFC-MER \cite{tricome_su2021keyfacial}, HTNet \cite{tricome_wang2024htnet}, and a task-adapted implementation of MMNet \cite{ijcai2022p150} on TriCoME. AlexNet provides a conventional convolutional baseline. KFC-MER uses first- and second-order motion from key facial components, while HTNet uses a hierarchical Transformer to represent facial movements. All four methods use only visual information.

For AlexNet, KFC-MER, and HTNet, we obtain the results by rerunning each method on TriCoME while retaining the training hyperparameters specified in its original paper, including the number of training epochs. MMNet is evaluated using the task-adapted implementation described above in HFFNet's \texttt{visual\_only} mode. Thus, all values in Table~\ref{tab:visual_baseline} come from experiments on TriCoME under method-specific training configurations.

The MMNet visual-only mode directly applies the adapted visual encoder and its three-class classifier. It bypasses the visual projection, sentence and context encoders, multimodal fusion, gating, and routing. The resulting prediction therefore depends only on visual input.

\begin{table}[!t]
\caption{Performance of Visual-Only Methods on TriCoME under Method-Specific Training Settings\label{tab:visual_baseline}}
\centering
\footnotesize
\setlength{\tabcolsep}{4pt}
\begin{tabular}{lcc}
\hline
Method & Accuracy (\%) & UF1 (\%) \\
\hline
AlexNet \cite{tricome_krizhevsky2012imagenet} & 66.12 & 51.40 \\
KFC-MER \cite{tricome_su2021keyfacial} & 69.11 & 62.32 \\
HTNet \cite{tricome_wang2024htnet} & 76.90 & 63.55 \\
MMNet (adapted) \cite{ijcai2022p150} & \textbf{82.96} & \textbf{77.87} \\
\hline
\end{tabular}

\vspace{2pt}
\begin{minipage}{\columnwidth}
\footnotesize
\textbf{Note:} AlexNet, KFC-MER, and HTNet are rerun on TriCoME with the training hyperparameters specified in their original papers, including epoch counts. MMNet (adapted) denotes the task-adapted implementation evaluated in HFFNet's visual-only mode.
\end{minipage}
\end{table}

Under the reported configurations, MMNet (adapted) has the highest reported Accuracy and UF1, at 82.96\% and 77.87\%, respectively. HTNet ranks second, with 76.90\% Accuracy and 63.55\% UF1. These results motivate the use of the adapted MMNet as HFFNet's visual backbone. Its scores are reused as the visual-only reference in the following ablation studies. The comparison characterizes the evaluated implementations under their respective training settings.

\subsection{Ablation Study}
We examine how context modeling, routing, and feature gating contribute to HFFNet on TriCoME, using the adapted MMNet as the visual-only reference. These experiments concern configurations of HFFNet, whereas Table~\ref{tab:visual_baseline} compares different visual methods under their respective training settings. Comparisons with the visual-only reference assess the complete fusion and classification pipeline, and the component ablations examine differences within that pipeline.

\subsubsection{Context Modeling and Routing}
Table~\ref{tab:overall_performance} compares five model configurations. CmT combines audio and text without visual input or contextual modeling. CmT+CoT adds the Contextual Transformer to incorporate dialogue history. HFFNet+LR and HFFNet+FR use all three modalities and differ in their routing threshold: LR learns the threshold, whereas FR fixes it at $\tau=0.8$. MMNet (adapted) is the same visual-only reference reported in Table~\ref{tab:visual_baseline}.

\begin{table}[!t]
\caption{Ablation of Context Modeling and Routing on TriCoME\label{tab:overall_performance}}
\centering
\begin{tabular}{lcc}
\hline
Method & Accuracy (\%) & UF1 (\%) \\
\hline
MMNet (adapted) & 82.96 & 77.87 \\
CmT & 70.36 & 58.62 \\
CmT+CoT & 79.63 & 70.83 \\
HFFNet+LR & 85.61 & 79.21 \\
HFFNet+FR & \textbf{85.89} & \textbf{83.26} \\
\hline
\end{tabular}
\end{table}

The adapted MMNet visual baseline reaches 82.96\% Accuracy and 77.87\% UF1, exceeding both audio-text configurations. Adding CoT to CmT increases Accuracy from 70.36\% to 79.63\% and UF1 from 58.62\% to 70.83\%, gains of 9.27 and 12.21 percentage points, respectively. Dialogue context improves the audio-text branch, but that branch still remains below MMNet (adapted) by 3.33 percentage points in Accuracy and 7.04 percentage points in UF1.

Within the complete framework, HFFNet+FR reaches 85.89\% Accuracy and 83.26\% UF1, compared with 85.61\% and 79.21\% for HFFNet+LR. Fixing the threshold therefore improves these metrics by 0.28 and 4.05 percentage points in the reported experiments. The UF1 difference concerns the mean of the class-wise scores; it does not by itself identify which emotion classes account for the gain.

\subsubsection{Gating and Routing}
We further evaluate gating and routing as a $2\times2$ ablation on top of the cross-modal attention baseline, denoted Base Fusion. Gating Only and Routing Only enable one component at a time, and Full HFFNet enables both. Visual-Only Baseline denotes the same adapted MMNet reported in Table~\ref{tab:visual_baseline}, and Naive Concatenation provides an additional reference configuration. Table~\ref{tab:ablation_fusion} reports the results.

\begin{table}[!t]
\caption{Fine-Grained Ablation Matrix of the Multimodal Fusion Module\label{tab:ablation_fusion}}
\centering
\begin{tabular}{lcc}
\hline
Module Configuration & Accuracy (\%) & UF1 (\%) \\
\hline
Visual-Only Baseline & 82.96 & 77.87 \\
Naive Concatenation & 74.37 & 70.55 \\
Base Fusion & 79.35 & 71.50 \\
Gating Only & 77.99 & 72.62 \\
Routing Only & 83.62 & 77.79 \\
Full HFFNet & \textbf{85.89} & \textbf{83.26} \\
\hline
\end{tabular}
\end{table}

Naive concatenation reduces Accuracy from the visual baseline's 82.96\% to 74.37\% and UF1 from 77.87\% to 70.55\%. Adding cross-modal attention raises the scores to 79.35\% Accuracy and 71.50\% UF1, but Base Fusion still falls below the visual-only result. Thus, adding audio and text without sufficient control over their contribution can reduce recognition performance on this dataset.

Gating Only reaches 77.99\% Accuracy and 72.62\% UF1. Its lower Accuracy than Base Fusion is consistent with useful information being suppressed along with irrelevant features, although its UF1 is slightly higher. Routing Only reaches 83.62\% Accuracy and 77.79\% UF1. Relative to the visual baseline, this is an Accuracy increase of 0.66 percentage points and a UF1 decrease of 0.08 percentage points. Routing therefore preserves much of the visual branch's performance but provides little improvement on its own.

Full HFFNet reaches 85.89\% Accuracy and 83.26\% UF1, exceeding either single-component configuration. Gating controls the sentence features passed to the multimodal classifier, while routing retains a visual-only decision for samples with high visual confidence. Their combined result supports using both controls in this framework.

\section{Discussion}

\subsection{Gating and Routing in Multimodal Fusion}
Our experiments show that additional modalities do not guarantee better recognition. Naive concatenation reaches 74.37\% Accuracy, below the adapted MMNet visual baseline at 82.96\%. Cross-modal attention raises Accuracy to 79.35\%, but still does not recover the visual-only result. As shown in Table~\ref{tab:ablation_fusion}, uncontrolled fusion can reduce performance on TriCoME. Irrelevant speech or text is one possible source of this interference.

The component ablations distinguish the roles of gating and routing. Gating alone yields 77.99\% Accuracy and 72.62\% UF1, both below the visual-only baseline. This suggests that suppressing sentence features can also remove useful information when the model has no visual-only fallback. Routing alone yields 83.62\% Accuracy and 77.79\% UF1. It keeps visual-only predictions for samples with $c_{vis}\geq\tau$, but sends low-confidence samples to a multimodal branch whose features have not been filtered by the gate. Combining the two components raises Accuracy to 85.89\% and UF1 to 83.26\%. Gating controls which sentence features enter fusion, while routing provides a choice between the visual-only and multimodal predictions.

\subsection{Fixed and Learnable Routing Thresholds}
The fixed-threshold variant reaches 83.26\% UF1, compared with 79.21\% for the learnable-threshold variant, a difference of 4.05 percentage points. The corresponding Accuracy scores are 85.89\% and 85.61\%. These results favor $\tau=0.8$ in the current TriCoME experiments. Fixing the threshold removes a routing parameter that would otherwise have to be estimated from a small training set, which may make routing less sensitive to sample variation. This provides a possible explanation for the observed difference, rather than evidence that a fixed threshold will always outperform a learned one.

\section{Conclusion}
We develop a brand-new dataset TriCoME and propose HFFNet for micro-expression recognition in real-world conversations. TriCoME pairs micro-expression videos with audio, transcripts, and annotations identifying the conversational role of preceding utterances. HFFNet combines fine-grained audio-text interaction with sequence compression, uses context-type embeddings to distinguish dialogue roles, and combines feature gating with confidence-based routing to control contextual interference. Together, TriCoME and HFFNet provide a data resource and a modeling framework for studying micro-expression recognition in natural conversations through visual, acoustic, and textual information coupled with role-annotated dialogue context. This work extends the conventional focus on isolated visual analysis in controlled laboratory settings toward the study of subtle facial behavior within real-world conversational interactions. Future work will focus on collecting larger multilingual and cross-cultural datasets to assess generalization across languages and cultures, and developing lightweight implementations of HFFNet for deployment on resource-constrained devices.

\section*{Acknowledgments}
This work was supported by the National Key R\&D Program of China under Grant 2023YFB2806900 and the Shenzhen Science and Technology Program under Grant KCXFZ20230731100700002.

\bibliographystyle{IEEEtran}
\bibliography{IEEEabrv,refs}

\newpage

\vfill

\end{document}